\documentclass[conference]{IEEEtran}

\IEEEoverridecommandlockouts
\usepackage{cite}
\usepackage{amsmath,amssymb,amsfonts}
\usepackage{amsthm}
\usepackage{algorithmic}
\usepackage{graphicx}
\usepackage{textcomp}
\usepackage{xcolor}
\usepackage{booktabs}
\usepackage{multirow}   % 提供 \multirow (跨行合并)
\usepackage{makecell}   % 提供 \makecell (单元格内换行/控制格式)
\makeatletter
\@ifpackageloaded{appendix}{\@namedef{ver@appendix.sty}{}}{}
\makeatother
\usepackage{hyperref}

\newtheorem{theorem}{Theorem}
\def\BibTeX{{\rm B\kern-.05em{\sc i\kern-.025em b}\kern-.08em
    T\kern-.1667em\lower.7ex\hbox{E}\kern-.125emX}}
\begin{document}

\title{From Path Signatures to Sequential Modeling: Incremental Signature Contributions for Offline RL\\
}

\author{
\IEEEauthorblockN{Ziyi Zhao\textsuperscript{*}\thanks{\textsuperscript{*}These authors contributed equally to this work.}}
\IEEEauthorblockA{\textit{TUM School of CIT}\\
\textit{Technical University of Munich}\\
Munich, Germany\\
zhaozy112@outlook.com}
\and
\IEEEauthorblockN{Qingchuan Li\textsuperscript{*}\footnotemark}
\IEEEauthorblockA{\textit{TUM School of CIT}\\
\textit{Technical University of Munich}\\
Munich, Germany\\
liqingchuan0119@gmail.com}
\and
\IEEEauthorblockN{Yuxuan Xu}
\IEEEauthorblockA{\textit{TUM School of CIT}\\
\textit{Technical University of Munich}\\
Munich, Germany\\
andreas.xu@tum.de}
}
\maketitle

\begin{abstract}
Path signatures embed trajectories into tensor algebra and constitute a universal, non-parametric representation of paths; however, in the standard form, they collapse temporal structure into a single global object, which limits their suitability for decision-making problems that require step-wise reactivity. We propose the Incremental Signature Contribution (ISC) method, which decomposes truncated path signatures into a temporally ordered sequence of elements in the tensor-algebra space, corresponding to incremental contributions induced by last path increments. This reconstruction preserves the algebraic structure and expressivity of signatures, while making their internal temporal evolution explicit, enabling processing signature-based representations via sequential modeling approaches. In contrast to full signatures, ISC is inherently sensitive to instantaneous trajectory updates, which is critical for sensitive and stability-requiring control dynamics. Building on this representation, we introduce ISC-Transformer (ISCT), an offline reinforcement learning model that integrates ISC into a standard Transformer architecture without further architectural modification. We evaluate ISCT on HalfCheetah, Walker2d, Hopper, and Maze2d, including settings with delayed rewards and downgraded datasets. The results demonstrate that ISC method provides a theoretically grounded and practically effective alternative to path processing for temporally sensitive control tasks.
\end{abstract}


\section{Introduction}
Path signature representation~\cite{lyons1998roughsignals} arise from rough path theory and provides a rigorous way to represent potentially irregular paths via collections of iterated integrals. The signature of a path forms a graded sequence of integrals that encode its geometric and ordered structure. Importantly, signatures depend only on the underlying geometric path up to monotone time reparameterization~\cite{lyonscaruana2007roughpaths}, which encourages a focus on the “shape” of a trajectory. This makes signatures particularly well suited for sequential data, as they can capture higher-order interactions and the ordering of events.

Several studies have demonstrated the effectiveness of path signatures in sequential modeling. Generally, as shown in~\cite{levin2013learning}, path signatures provide a rich, non-linear feature representation of paths, under which continuous path-dependent functions can be approximated by linear models acting on the signature. Building upon this representation, log-signatures further remove algebraic redundancies while preserving expressive power, leading to more compact representations~\cite{morrill2021neuralrde}. Beyond fixed feature extraction in traditional sense, signature-based representations have also been successfully integrated into learnable models: for instance, \cite{graham2013sparse} and \cite{tong2023sigformer} apply signatures to handwriting recognition and financial time series, respectively, demonstrating the stability and scalability of signature in modeling irregular sequences with interacting dimensions.

When sequential modeling is extended to continuous control, the learning problem strongly demands a model’s ability to extract structured, inter-dimensional dependencies from historical data. Many real-world decision processes align better with Partially Observable Markov Decision Process (POMDP) settings, where single-step observations are insufficient as complete statistical measures of system dynamics. In such problems, policies must effectively utilize history to resolve observation ambiguity and perform long-term credit allocation. In recent years, the Decision Transformer (DT)~\cite{chen2021dt} and its variants~\cite{furuta2022generalizeddecisiontransformeroffline}~\cite{zheng2022onlinedecisiontransformer}~\cite{correia2022hierarchicaldecisiontransformer} have modeled reinforcement learning as a conditional sequence modeling task, thereby avoiding some instability factors of traditional bootstrapping approaches and achieving competitive performance on offline RL benchmarks. However, in DT-like architectures, only one simple embedding projection per token is used to capture the inter-dimensional information. Although subsequent attention and feed-forward layers are highly non-linear, the initial token-level projection compresses high-dimensional observations to the latent space in a point-wise manner, without explicitly encoding temporal higher-order cross-dimensional interactions, placing the burden of recovering such structure entirely on downstream attention mechanisms. On the other hand, while recurrent structures like RNNs offer more compact historical compression~\cite{ni22a}, they remain constrained by vanishing gradients and finite memory capacity on long sequences, struggling to capture high-order, long-range dependencies.

Against this backdrop, this paper focuses on the role of path signature representations in offline decision learning. In prior work applying path signatures to reinforcement learning, signature representations are often used in a relatively coarse manner. For example, in SigFormer~\cite{tong2023sigformer}, truncated path signatures of order 
$k$ are aggregated into one or two tokens and directly fed into a Transformer architecture. While this design preserves the complete set of $k$-th order signature information over the entire trajectory, it does not provide a fine-grained characterization of the temporal evolution of signature contributions.

As the trajectory length increases, the incremental influence of recent observations on the global signature diminishes, reducing sensitivity to local or abrupt state changes. In Section \ref{sec:Ablation}, our ablation results further show that models relying on full-path signature tokens exhibit inferior sensitivity to local dynamics compared to incremental signature representations, leading to degraded performance in continuous control tasks.

This motivates a representation that preserves the expressive power of path signatures while maintaining sensitivity to local temporal dynamics. Our contribution focuses on:

\begin{itemize}
    \item We propose the Incremental Signature Contribution (ISC) method, which decomposes a global path signature into a sequence of time-indexed incremental components, explicitly characterizing the contribution of each observation to the overall signature. This construction alleviates the limitations of full-path signature representations in capturing local and abrupt dynamical changes.
    \item We introduce the Incremental Signature Contribution Transformer (ISCT), an offline reinforcement learning model built upon ISC representations. ISCT achieves competitive performance on continuous control and navigation benchmarks, and exhibit strong robustness under delayed reward settings. 
\end{itemize}

Existing approaches leverage path signatures in different ways, ranging from local window-based signature features to log-signature–driven Neural Rough Differential Equations (Neural RDEs) that update hidden states via block-wise ODE flows. In contrast, our method exposes the incremental accumulation of signature information as a time-indexed sequence, enabling step-wise modeling of path geometry while preserving algebraic consistency through the Chen identity.

To support reproducibility, we provide codes in this \href{https://github.com/lunatic112/Incremental-Signature-Contribution}{repository}.

\section{related work}\label{sec:2}

\subsection{Offline reinforcement learning}
Offline Reinforcement Learning (RL) aims to extract optimal control policies from static, pre-collected transition datasets without active exploration. The primary hurdle in this paradigm is distributional shift, where the learned policy queries state–action pairs outside the training distribution, often leading to extrapolation error and catastrophic value overestimation. To maintain stability, a foundational class of methods employs policy constraints and regularization. Representative algorithms like CQL~\cite{kumar2020cql} and TD3+BC~\cite{fujimoto2021td3bc} penalize out-of-distribution (OOD) actions, while Implicit Q-Learning (IQL)~\cite{kostrikov2022iql} mitigates overestimation by framing offline learning as an in-sample dynamic programming problem. Parallel to these value-based approaches, a growing line of work leverages generative modeling to capture multi-modal behavior in demonstrations, notably diffusion-based policy learning~\cite{wang2022diffusionql}~\cite{chi2023diffusionpolicy}. Another influential paradigm formulates offline RL as conditional sequence modeling, where a Transformer is trained to predict actions conditioned on past trajectories and desired returns. Decision Transformer (DT)~\cite{chen2021dt} and subsequent variants such as Trajectory Transformer~\cite{janner2021trajectorytransformer} have demonstrated competitive performance on a range of offline RL benchmarks while sidestepping some instabilities of bootstrapping-based methods.

\subsection{Path signature in neural network}
In recent years, path signatures have evolved from “fixed features” to “end-to-end trainable network modules” as hierarchical iterative integral representations of sequences, enabling the construction of stronger temporal inductive biases. Classic Deep Signature Transforms treat signatures as differentiable pooling/layers that can be inserted into networks, enabling data-dependent feature selection through learnable augmentations. In continuous-time modeling, Neural CDE treats irregularly sampled sequences as control paths to drive hidden state evolution~\cite{kidger2020ncde}, while Neural RDE further employs local log-signature summaries of long sequences to drive effects, enhancing scalability~\cite{morrill2021neuralrde}. Recent work continues along this trajectory: examples include Learnable Path NCDE for optimized input path representations and Log-NCDE based on Log-ODE/Li bracket structures, improving efficiency and effectiveness on longer sequences and more complex datasets~\cite{jhin2023learnablepath,walker2024logncde}. On the other hand, signatures have been integrated into modern attention mechanisms: Rough Transformers~\cite{moreno2024roughtransformer} enhance attention through “signature patching / multi-view signature attention” to handle irregular and long-dependency sequences in continuous time; SigFormer~\cite{tong2023sigformer} combines signatures with Transformers for sequence decision tasks like deep hedging. In generative modeling, SigDiffusions~\cite{barancikova2024sigdiffusions} transfers score-based diffusion to log-signature embedding spaces for generating multivariate time series, reporting structural preservation and performance. Beyond these approaches, some researchers~\cite{zhao2024cmfil} have integrated signatures into model inputs for online reinforcement learning to optimize policies. 

However, these efforts have not extended into offline reinforcement learning. Therefore, this paper attempts to introduce signature as a control feature in offline-RL, further refining their application in reinforcement learning-related domains.

\section{Preliminary}

\subsection{Path Signature}

Path signature is a powerful feature extraction technique for sequential data, and has been widely applied in multivariate time series analysis.
It provides a systematic way to characterize the geometric and temporal structure of a path through a collection of iterated integrals.

Let a path be defined as
\begin{equation}
X = \{x_t \mid x_t = (x_t^1,\ldots,x_t^d) \in \mathbb{R}^d,\; t \in [a,b]\},
\end{equation}
where $a,b \in \mathbb{R}$ denote the time interval and $d$ is the dimension of the signal.
The signature of the path in the Stratonovich sense is defined as
\begin{equation}
\mathrm{Sig}(X) = \left(1,\; \mathrm{Sig}_1(X),\; \mathrm{Sig}_2(X),\; \ldots,\; \mathrm{Sig}_i(X),\; \ldots \right),
\end{equation}
where the $i$-th level signature is given by
\begin{equation}
\mathrm{Sig}_i(X)
=
\int_{a<t_1<\cdots<t_i<b}
\circ d x_{t_1} \otimes \cdots \otimes \circ d x_{t_i},
\quad i=1,2,\ldots
\end{equation}
Here, $\circ$ denotes the Stratonovich integral and $\otimes$ denotes the tensor product.
By convention, the zeroth-level term is defined as $\mathrm{Sig}_0(X)=1$.

The first-order signature captures the overall displacement of the path.
Higher-order terms encode increasingly complex temporal patterns, including curvature and cross-channel interactions.
Under mild conditions, the signature uniquely characterizes the path up to tree-like equivalence, making it a faithful representation of sequential data.

In practice, the data in the common reinforcement datasets are always discretized, so we can only obtain numerical values at a finite set of discrete time points.
Hence, we need the discretized form of path $X(t)$ on the interval $[a,b]$ at time points
\[
a=t_0<t_1<\cdots<t_N=b,\qquad x_n := X(t_n),
\]
and define the increments on each subinterval by
\[
\Delta x_n := x_{n+1}-x_n \in \mathbb{R}^d,\qquad n=0,\ldots,N-1.
\]
Then a discrete approximation of the $i$-th level signature is given by
\[
\mathrm{Sig}^{\Delta}_i(X)
\;:=\;
\sum_{0\le n_1<n_2<\cdots<n_i\le N-1}
\Delta x_{n_1}\otimes \Delta x_{n_2}\otimes\cdots\otimes \Delta x_{n_i}.
\]

Similarly, the full signature is infinite-dimensional and therefore intractable.
A truncated signature of order $N$ is commonly used:
\begin{equation}
\mathrm{Sig}^N(X) = \left(1,\; \mathrm{Sig}_1(X),\; \ldots,\; \mathrm{Sig}_N(X)\right),
\end{equation}
where $N \in \mathbb{N}^+$ denotes the truncation level.
This truncation provides a trade-off between expressiveness and computational efficiency.

Path signature is closely related to rough path theory, which establishes a rigorous mathematical framework for analyzing irregular signals.
By augmenting a path with its iterated integrals, rough path theory enables stable modeling and differential equation analysis for highly non-smooth trajectories.

\subsection{Universal Nonlinearity Approximation}

\begin{theorem}[\textbf{Factorial Decay} (Theorem 3.2 in~\cite{lyons2025signaturemethodsmachinelearning})]
Let $V$ be a Banach space, $[a,b] \subset \mathbb{R}$ a compact interval, and $\gamma \in \mathcal{V}^1([a,b], V)$ a path of bounded variation. Let $S(\gamma) : \Delta_{[a,b]} \to T((V))$ denote the signature of $\gamma$. Then for every $n \in \mathbb{Z}_{\geq 1}$ and every $(s,t) \in \Delta_{[a,b]}$, the $n$-th component of the signature $S_n(\gamma) : \Delta_{[a,b]} \to V^{\otimes n}$ satisfies:
\begin{equation}
    \|S_{s,t}^n(\gamma)\|_{V^{\otimes n}} \leq \frac{\|\gamma\|_{1,[a,b]}^n}{n!}
\end{equation}
where $\|\gamma\|_{1,[a,b]}$ denotes the total variation of the path $\gamma$ on the interval $[a,b]$.
\end{theorem}

\begin{theorem}[Universal Nonlinearity (Theorem 3.3 in~\cite{lyons2025signaturemethodsmachinelearning})]
For any compact set $K \subset \mathrm{Path}^1([a,b],V)$ and any continuous function 
$F:K\rightarrow\mathbb{R}$, for any $\varepsilon>0$, there exists a truncation level $n$ and real coefficients
$\{\alpha_i(J)\}$ such that
\begin{equation}
\Big|
F([\theta]) -
\sum_{i=0}^{n}
\sum_{J\in\{1,\ldots,d\}^i}
\alpha_i(J) S_{a,b}(\theta)^J
\Big|
\le \varepsilon.
\end{equation}
\end{theorem}

The mathematical property of these two theorems provides the foundation for our offline reinforcement learning Framework:
\begin{itemize}
    \item \textbf{Justification for Truncation:} In a norm-based sense, the truncation of the signature to a finite depth $K_0$ manages to capture the terms with the most significant contribution. Due to the $n!$ term in the denominator, higher-order terms disappear rapidly. Thus, a sufficiently large $K_0$ effectively summarizes the path's geometry.
    
    \item \textbf{Universal Nonlinearity:} The framework approximates a system's response via a linear combination of signature components. The signature acts as a universal nonlinearity on paths: complex continuous functions of paths behave approximately linearly when mapped into the high-dimensional signature space.
\end{itemize}
In general, the above theorems conclude that, up to an appropriate truncation level, the path signature can approximate any function $F\in K\rightarrow\mathbb{R}$ with any precision via a linear transformation.

\section{Methods}

\subsection{Incremental Signature Contribution}
Under a discrete time setting as above, we introduce the computation of our ISC method. Same as path signatures, ISC is also defined in infinitely many levels. In particular, for time step $t_n$, ISC at that time is defined as
\begin{equation}
    \Delta S_{n} := \{\Delta S^{(1)}_{n}, \Delta S^{(2)}_{n}, \Delta S^{(3)}_{n}...\}
\end{equation}
For level 1, the ISC is computed as:
\begin{equation}
    \Delta S^{(1)}_{n} = \Delta x_{n}
\end{equation}

For level 2, it is computed as:
\begin{equation}
    \Delta S^{(2)}_{n} = S^{(1)}_{n-1} \otimes \Delta x_{n} + \frac{1}{2} \Delta x_{n} \otimes \Delta x_{n},
\end{equation}
where $S^{(1)}_{n-1}$ denotes the first-order signature until time $t_{n-1}$.

More generally, we can extend the above incremental construction to an $n$-th order truncation.
Let $S^{(k)}_{n-1}$ denote the \emph{$k$-th order} signature state up to time $n-1$ (with the convention
$S^{(0)}_{n-1}=1$). For each $k=1,\dots,n$, we define the incremental $k$-th order signature feature as
\begin{equation}
\Delta S^{(k)}_{n}
=\sum_{j=1}^{k}\frac{1}{j!}\; S^{(k-j)}_{n-1}\otimes\big(\Delta x_{n}\big)^{\otimes j},
\qquad k=1,\ldots,n,
\label{eq:delta_sig_k}
\end{equation}
where $(\Delta x_{n})^{\otimes j}$ denotes the $j$-fold tensor product of $\Delta x_{n}$.
Equivalently, expanding the upper expression, it can be written as
\begin{equation}
\Delta S^{(k)}_{n}
= S^{(k-1)}_{n-1}\otimes \Delta x_{n}
+ \frac{1}{2} S^{(k-2)}_{n-1}\otimes (\Delta x_{n})^{\otimes 2}
+\cdots
+\frac{1}{k!}(\Delta x_{n})^{\otimes k}.
\end{equation}
The resulting tensors are flattened and fed to embedding projections, specially, we name the first level ISC as INC token and the second as CROSS token.

It is worth noting that the full path signature up to time $t_N$ can be exactly recovered from the sequence of Incremental Signature Contributions (ISC). This follows directly from Chen’s identity, which states that the signature of a path over a time interval can be factorized as the tensor product of signatures over its sub-intervals~\cite{lyonscaruana2007roughpaths}. Consequently, the proposed ISC decomposition preserves the complete information encoded in the original path signature, while reorganizing it into a temporally structured representation that is amenable to sequential modeling.

\subsection{Incremental Signature Contribution Transformer}

\begin{figure}[t]
    \centering
    \includegraphics[width=1.0\linewidth]{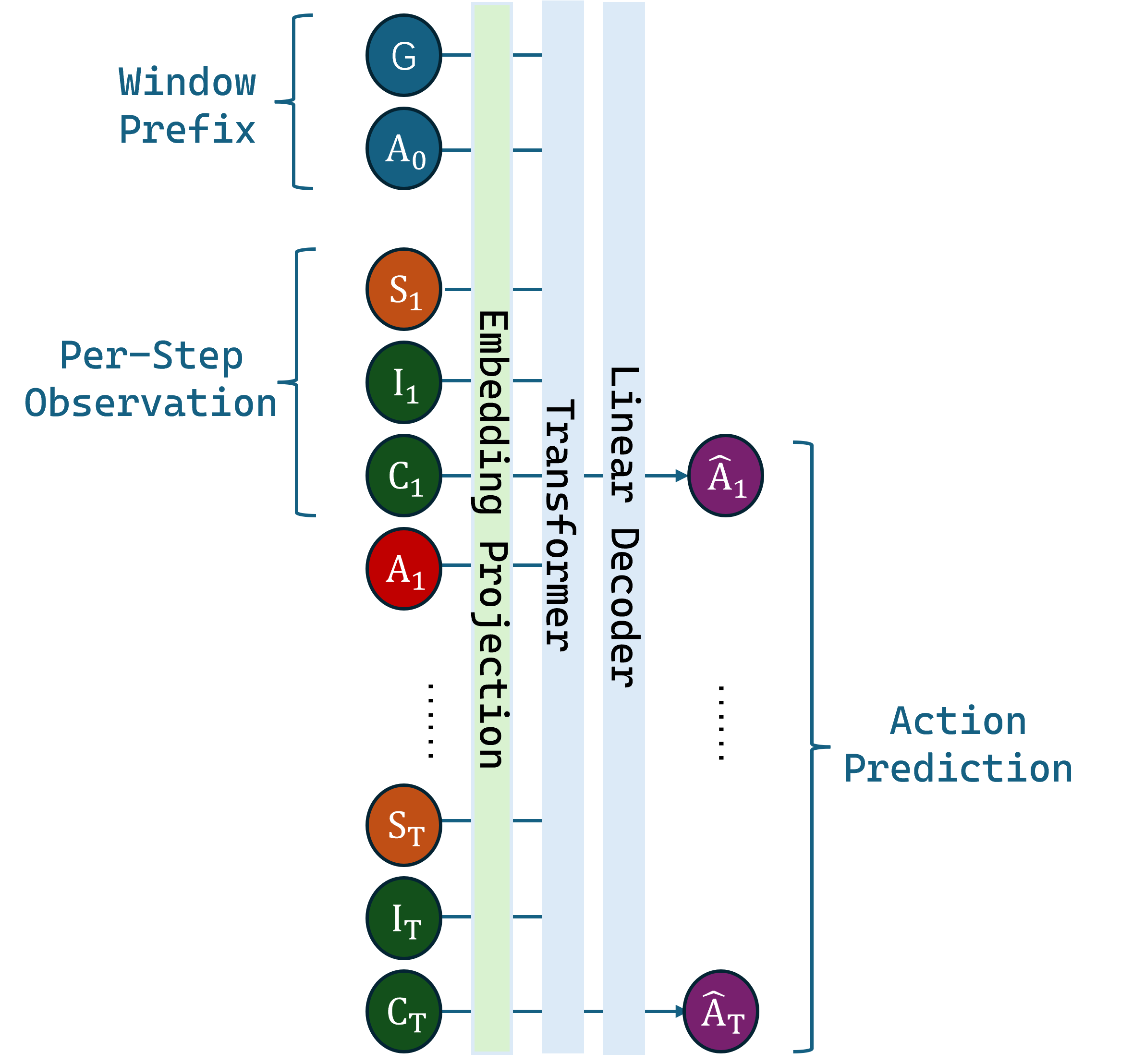}
    \caption{Structure of ISCT.$G$ denotes goal token, $A_n$, $S_n$, $I_n$, $C_n$, $\hat{A}_n$ represents the action, state, INC, CROSS, and action prediction token on time step $t_n$, respectively. We use the convention that $t_0$ denotes the last step before the observed window.}
    \label{fig:structure}
\end{figure}

By applying ISC to encode trajectories into token sequences for transformer-based decision making(shown in Fig.\ref{fig:structure}), we propose a model named ISCT for offline reinforcement learning. Similar to DT, our model observes only a fixed length window in data stream. But instead of using raw state-action pairs, we add the first and the second level of ISC in window building, providing supplementary information. We also replace return-to-go, often used in the DT ans its variants, by a simple Goal-token, which is computed by summing all rewards inside a window. Moreover, we do not include any time-information in our model.

For each training window of length $T$, we construct the input sequence as:
\begin{equation}
[G,\; a_{n_0-1},\; \{ \text{OBS}(x_n),\;
\text{INC}(\Delta S^{(1)}_{n}),\;
\text{CROSS}_g(\Delta S^{(2)}_{n}) \}_{t=0}^{N}],
\end{equation}

where $G$ denotes the goal token and $a_{n_0-1}$ is the previous action. 

In practice, to ease computation burden resulting from exponentially increasing dimension of ISC, we may divide the increments $\Delta x_t$ into $C$ many channels. For each channel, we compute ISC separately, and concatenate them together before fed to the network.

All token types are projected into a shared latent space through type-specific linear embedding layers. We further add token-type, channel, and position embeddings to preserve structural information. Since signature is invariant to monotone reparameterization, position embedding already provides enough temporal constraint for the model. This design enables our structure ISCT to predict jointly over absolute states, local dynamics, and higher-order temporal patterns, while being inheretedly robust to delayed rewards and uneven observation sampling.

\subsection{Comparison with Other Signature Usage}

The construction above highlights a distinct mode of utilizing path signatures. In prior work, local signature approaches compute window-level signatures as fixed features that do not directly participate in temporal state evolution. Log-signature–based Neural Rough Differential Equations (Neural RDEs) instead summarize entire path segments and drive hidden-state updates via block-wise ODE flows, thereby treating the signature as an indivisible interval-level object.

By contrast, our Incremental Signature Contribution (ISC) method, while derived directly from the Chen identity and introducing no new algebraic objects, reorganizes the signature into temporally ordered incremental contributions within the tensor algebra. This makes explicit how signature information accumulates along the trajectory and enables its use as a time-indexed representation in standard sequential models.

\section{Experiment}

\subsection{General Results}
We perform our methods ISCT on four benchmark datasets in our experiments. (Shown in Table \ref{Table1})
We leverage datasets across several domains in Gym tests~\cite{fu2020d4rl}. For locomotive control problems, we experimented on halfcheetah, hopper and walker2d. Every offline datasets are generated by three collecting strategy (medium, medium-replay and medium-expert). Besides, we also tested our network on navigation task dataset Maze2d, in which the model aims to control a ball from a random starting position to a predefined target. We employ this method on three different sizes of mazes including umaze, medium-maze and large-maze.

Regarding the maze problem, we believe that beyond measuring the relevant scores, it is more crucial to assess the quality of the paths. In our training set for the Maze2d tasks, the starting and goal positions are randomized rather than fixed. To consistently evaluate and visualize the learned trajectories during the demonstration phase, we employ a fixed 'farthest-point' setup: for the U-maze, the start and end points are set at the two opposite tips of the 'U' shape; for the Medium and Large mazes, we designate diagonally opposite corners as the starting and ending positions. Therefore, we select a specific starting point and plot the distance to the destination to demonstrate its performance. We have also incorporated the path length required for movement into the evaluation criteria.

\begin{table}[t]
\centering
\small
\caption{Performance in gym tasks. Performance for ISCT comes from average and standard deviation scores are reported over 5 seeds and 10 episodes. Other results are reported by~\cite{kostrikov2022iql} and \cite{fujimoto2021td3bc}.}
\begin{tabular}{lccccc}
\toprule
\textbf{Environment} & \textbf{TD3+BC} & \textbf{CQL} & \textbf{IQL} & \textbf{DT} & \textbf{ISCT} \\
\midrule
% --- HalfCheetah Group ---
        halfcheetah-m   & 42.8  & 44.0  & \textbf{47.4}  & 42.6  & 42.9$\pm$1.1 \\
        halfcheetah-m-r & 43.3  & \textbf{45.5}  & 44.2  & 36.6  & 41.0$\pm$0.9 \\
        halfcheetah-m-e & \textbf{97.9}  & \textbf{91.6}  & 86.7  & 86.8  & \textbf{91.4}$\pm$ \textbf{1.5} \\
\midrule
% --- Hopper Group ---
        hopper-m       & \textbf{99.5}  & 58.5  & 66.3  & 67.6  & 58.8$\pm$7.2 \\
        hopper-m-r     & 31.4  & \textbf{95.0}  & \textbf{94.7}  & 82.7  & 70.5$\pm$9.0 \\
        hopper-m-e     & \textbf{112.2} & 105.4 & 91.5  & 107.6 & \textbf{109.8$\pm$6.4} \\
\midrule
% --- Walker2d Group ---
        walker2d-m     & \textbf{79.7}  & 72.5  & \textbf{78.3}  & 74.0  & \textbf{79.8$\pm$8.2} \\
        walker2d-m-r   & 25.2  & \textbf{77.2}  & 73.9  & 66.6  & 68.9$\pm$17.9 \\
        walker2d-m-e   & 101.1 & 108.8 & \textbf{109.6} & 108.1 & \textbf{109.1$\pm$0.5} \\
\midrule
\textbf{Average}       & \textbf{70.3} & \textbf{77.6} & \textbf{77.0} & \textbf{74.7} & \textbf{74.7} \\
\midrule[1pt] 

% --- Maze2d Group ---
        maze2d-u       & --    & \textbf{72.7}  & --    & -6.8  & 60.2$\pm$0.5 \\
        maze2d-m       & --    & 70.9  & --    & 31.5  & \textbf{86.8$\pm$1.2} \\
        maze2d-l       & --    & 90.9  & --    & 45.3  & \textbf{112.5$\pm$0.5} \\
\midrule
\textbf{Maze2d-avg}    & --    & \textbf{78.2} & --    & \textbf{23.3} & \textbf{86.5} \\
\bottomrule
\end{tabular}
\label{Table1}
\end{table}

Figures 2-4 demonstrate that path signatures are highly beneficial for navigation problems. By incorporating signature representations, the model becomes more sensitive to the temporal ordering of events, enabling the agent to better anticipate when to make turns and to reach the target more efficiently.

\begin{figure}[t]
    \centering
    \includegraphics[width=1\linewidth]{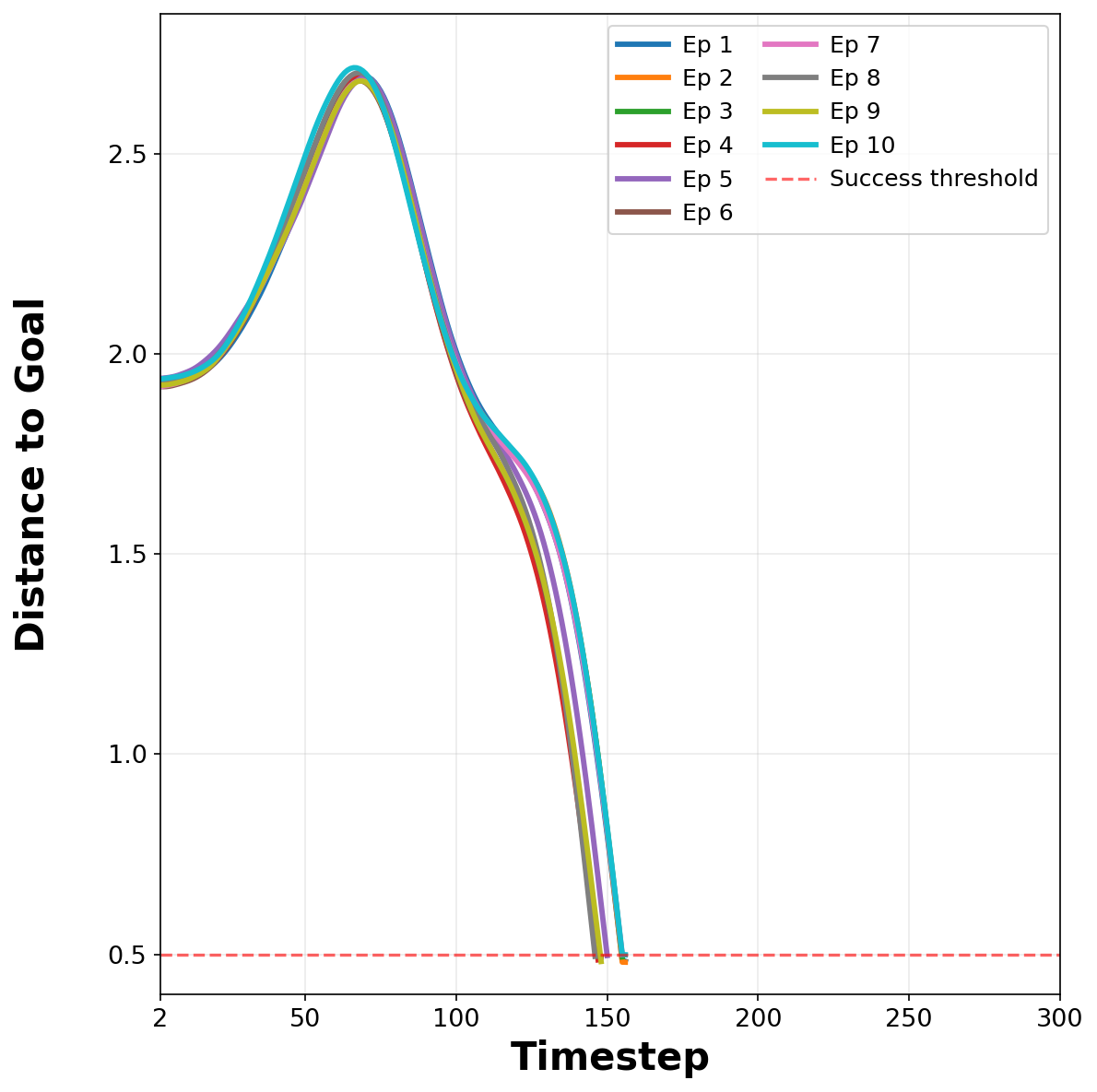}
    \caption{Path length comparison on U-Maze tasks.}
    \label{fig:umaze.png}
\end{figure}

\begin{figure}[t]
    \centering
    \includegraphics[width=1\linewidth]{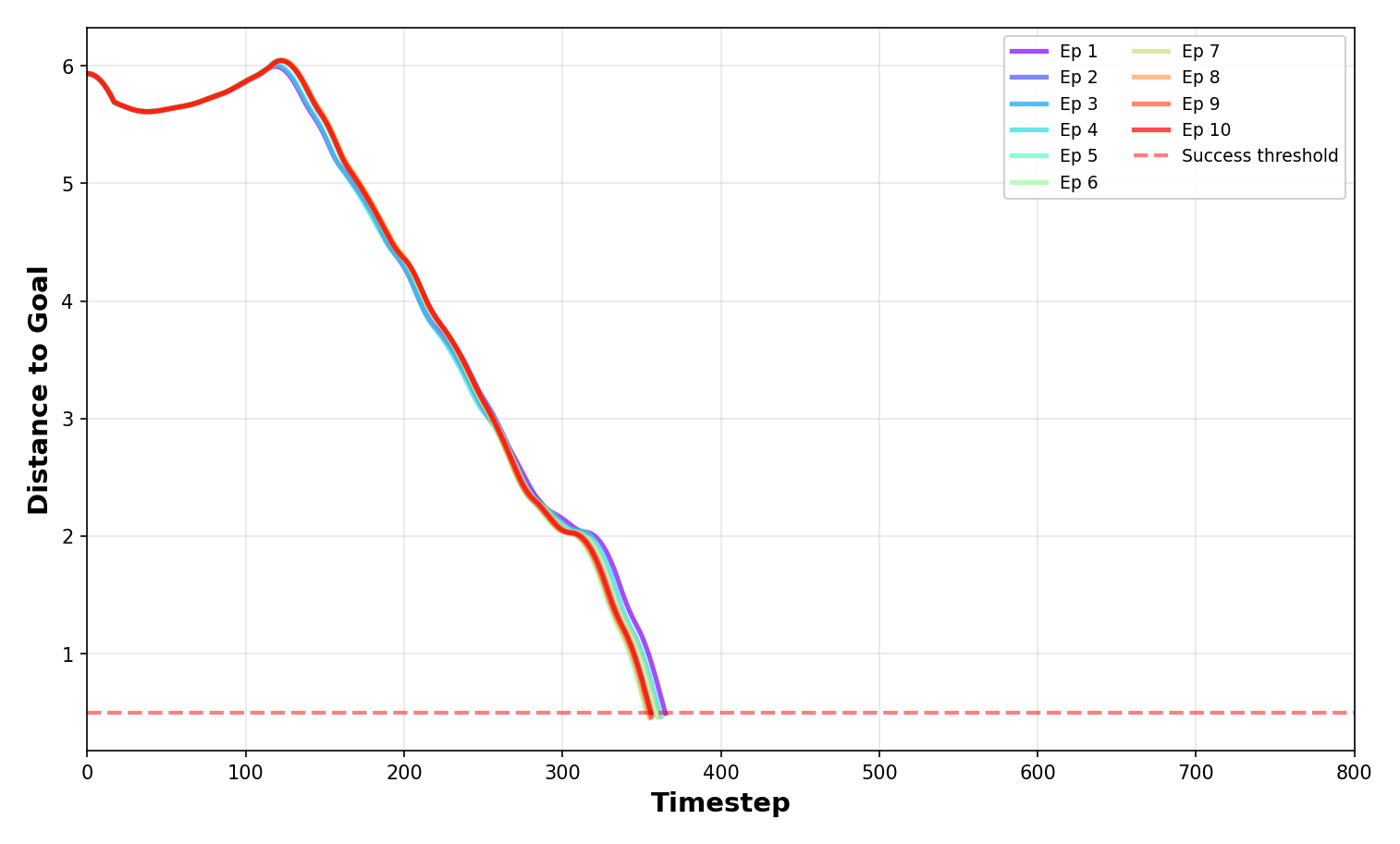}
    \caption{Path length comparison on Middle size Maze tasks.}
    \label{fig:mmaze}
\end{figure}

\begin{figure}[t]
    \centering
    \includegraphics[width=1\linewidth]{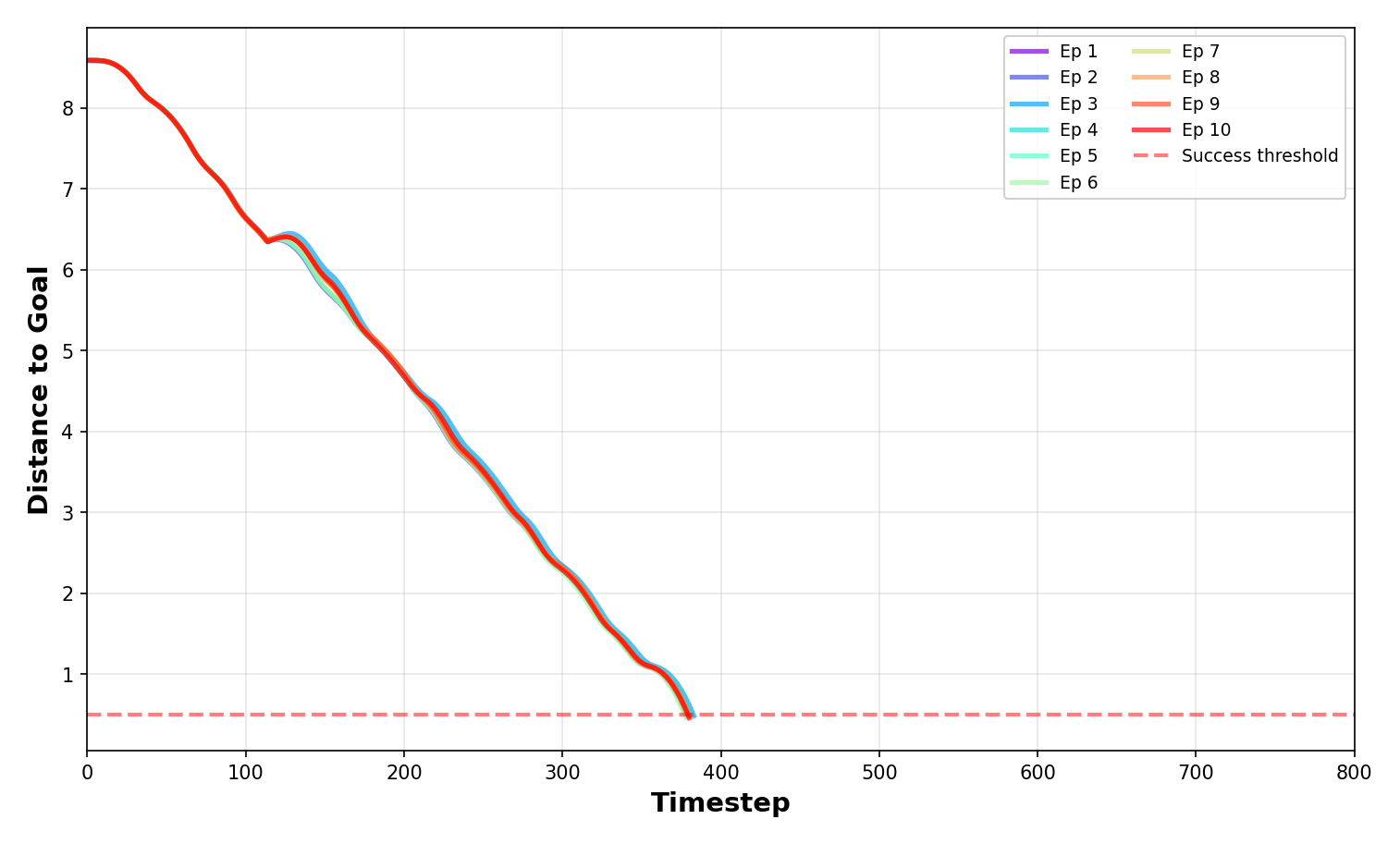}
    \caption{Path length comparison on Large size Maze tasks.}
    \label{fig:lmaze}
\end{figure}

\subsection{Gym Task with Delayed Rewards}
In this subsection, we investigate the performance of ISCT with delayed rewards in Mujoco locomotive task, i.e., under the constraint that the agent may receive the total reward only after termination, and also can only see the total returns of the trajectories in the training set. This is a practical situation to test, as in reality, per-step reward is not always available. We test on HalfCheetah dataset, and compare with CQL and DT, the performance is displayed in \ref{DelayedReward}. The results of CQL and DT comes from \cite{yamagata2023qlearningdecisiontransformerleveraging}, and the result from our model is computed through averaging total return on all windows across the trajectory. From the table, we can see that our model is virtually unaffected by delayed rewards, which is actually a foreseeable phenomenon, as our method, in its design, does not rely on per-step reward. The window goal in our method only classifies the quality of trajectory, but provide no information on pattern recognition or the state of the agent. By contrast, any Q-Learning method requires per-step reward to compute optimal value function, and DT-style methods also need per-step reward to update return-to-go. As shown in the table, both methods are heavily affected by delayed rewards. Hence, our method provides an alternative in offline-learning, which is built to be compatible with delay reward.

\begin{table}[t]
\centering
\small
\caption{Performance on Halfcheetah with delayed reward. Average and standard deviation scores are reported over 3 seeds and 10 episodes.}
\begin{tabular}{lcccc}
\toprule
\textbf{Environment} & \textbf{CQL} & \textbf{DT} & \textbf{QDT} & \textbf{ISCT} \\
\midrule
% --- Medium Group ---
halfcheetah-m  & 1.0  & 42.2          & 42.4 & \textbf{42.9} \\
% --- Medium Replay Group ---
halfcheetah-m-r& 7.8  & 33.0 & 32.8 & \textbf{42.4} \\
\bottomrule
\end{tabular}
\label{DelayedReward}
\end{table}

\subsection{Downgraded Training Set}
\begin{figure}[t]
    \centering
    \includegraphics[width=1\linewidth]{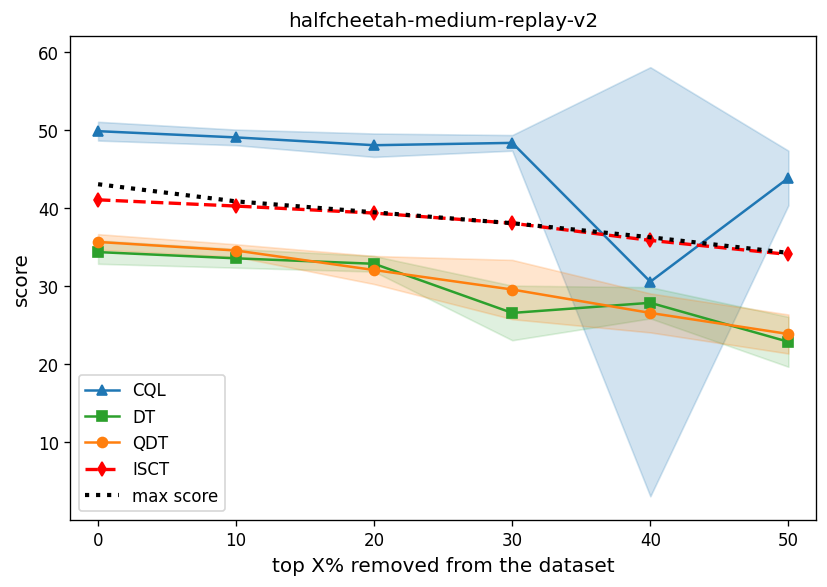}
    \caption{Performance on Downgraded Training Set}
    \label{fig:cut}
\end{figure}

In this subsection, we evaluate the performance of our ISCT model on the downgraded halfcheetah-medium-replay dataset to demonstrate its robustness. We delete the $X\%$   best trajectories from the dataset, where $X\in\{10, 20, 30, 40, 50\}$, and train our model on it. We plot the results in Figure \ref{fig:cut}, against the results of DT, CQL and QDT from \cite{yamagata2023qlearningdecisiontransformerleveraging}. Generally,  our model outperforms DT, which has the same transformer structure, and keeps tracking the best performance in the downgraded dataset. This shows the capability of our model in learning with suboptimal data.

\section{Ablation}\label{sec:Ablation}

\begin{table}[t]
\centering
\caption{Ablation on Increment Representation Strategy. Average and standard deviation scores are reported over 3 seeds and 10 episodes.}
\label{tab:ablation_representation}
\small
\renewcommand{\arraystretch}{1.2} 
\begin{tabular}{l c c}
\toprule
\textbf{Environment} & \textbf{Signature} & \textbf{No Signature} \\
\midrule

% --- halfcheetah-m ---
halfcheetah-m   & 42.9$\pm$1.1 & 42.3$\pm$0.9 \\

% --- hopper-m ---
halfcheetah-m-r      & 41.0$\pm$0.9 & 40.7$\pm$0.6 \\

% --- walker2d-m ---
halfcheetah-m-e     & 91.4$\pm$1.5 &   41.8$\pm$1.2\\
\midrule

% --- halfcheetah-m-e ---
walker2d-m & 77.3$\pm$8.2 & 60.5$\pm$20.1 \\

% --- hopper-m-e ---
walker2d-m-r     & 66.7$\pm$17.9 & 47.6$\pm$25.0 \\

% --- walker2d-m-e ---
walker2d-m-e    & 107.6$\pm$0.5 & 74.3$\pm$16.9 \\

\midrule[1pt] 
% --- 平均值 ---
\textbf{avg}    & 71.2 & 51.2 \\
\midrule[1pt] 

% --- maze2d 系列 ---
maze2d-u         & 60.2$\pm$0.5& 20.5$\pm$2.1\\

maze2d-m         & 86.8$\pm$1.2& 18.3$\pm$5.2 \\

maze2d-l         & 112.5$\pm$0.5& 30.5$\pm$10.5 \\
\bottomrule
\end{tabular}
\label{table3}
\end{table}

\begin{table}[t]
\centering
\caption{Ablation on Signature Injection Strategy. Average and standard deviation scores are reported over 3 seeds and 10 episodes.}
\label{tab:ablation_representation}
\small
\renewcommand{\arraystretch}{1.2} 
\begin{tabular}{l c c}
\toprule
\textbf{Environment} & \textbf{ISCT} & \textbf{Full signature} \\
\midrule

% --- halfcheetah-m ---
halfcheetah-m   & 42.9$\pm$1.1 & 29.9$\pm$11.5 \\

% --- hopper-m ---
halfcheetah-m-r      & 41.0$\pm$0.9 & 22.9$\pm$12.9 \\

% --- walker2d-m ---
halfcheetah-m-e     & 91.4$\pm$1.5 & 10.0$\pm$3.5 \\
\midrule

% --- halfcheetah-m-e ---
walker2d-m & 77.3$\pm$8.2 & 73.7$\pm$10.1 \\

% --- hopper-m-e ---
walker2d-m-r     & 66.7$\pm$17.9 & -0.9$\pm$0.1 \\

% --- walker2d-m-e ---
walker2d-m-e    & 107.6$\pm$0.5 & 94.7$\pm$21.4 \\

\midrule[1pt] 
% --- 平均值 ---
\textbf{avg}    & 71.2 & 38.3 \\
\bottomrule
\end{tabular}
\label{table4}
\end{table}

\subsection{Ablation on Increment Representation}
In this ablation study, we investigate the performance of ISCT when signature-based representations are removed, with the goal of isolating the contribution of path signatures to the overall model. To compensate for the structural change induced by removing ISC tokens—specifically, the reduced window length—we replace each ISC token with correlation-based features of the corresponding order. Correlations are chosen because they match the dimensionality of ISC at each level, thereby keeping the number of model parameters unchanged. Moreover, all correlation features are normalized to the same scale as the ISC tokens, ensuring that the comparison isolates the effect of representational content rather than magnitude or capacity differences. As shown in Table~\ref{tab:ablation_representation}, substituting ISC tokens with correlation-based features leads to a substantial degradation in performance, highlighting the critical role of incremental signature representations in ISCT.

\subsection{Ablation on Signature Injection Strategy}
In this experiment(shown in Table \ref{table4}), we removed the relevant increment signature tokens from ISCT and instead adopted a signature representation similar to other related works: injecting the entire truncated signature information into a single token. Under this setting, the model still functioned, but we observed a noticeable performance decline. This aligns with our intuition: indiscriminately injecting the full signature information not only reduces the model's ability to identify critical changes but also introduces redundant information to some extent. In contrast, our approach explicitly introduces incremental signature representations at each time step, enabling the model to capture state changes more precisely and respond more sensitively to dynamic transitions. Simultaneously, this incrementally updated representation proves more conducive to stimulating the attention mechanism for effectively modeling temporal variations than the one-time injection of full information.

\section{Conclusion and Future Work}
This paper introduced the Incremental Signature Contribution (ISC) framework, which enables sequential modeling directly on path signatures by decomposing a global signature into time-indexed incremental components, rather than treating it as a single static object. Building on this representation, we proposed the Incremental Signature Contribution Transformer (ISCT), an offline reinforcement learning model that applies a standard Transformer architecture to ISC tokens. Empirically, ISCT achieves competitive performance across continuous control and navigation benchmarks, while remaining robust in settings without per-step reward signals during both training and evaluation. These results suggest that explicitly modeling the nonlinear structure of path-dependent functionals at the representation level can substantially ease optimization and improve learning effectiveness in offline decision-making problems.

Future research directions span several promising avenues. First, we aim to extend ISC to longer-horizon and real-world tasks by investigating adaptive truncation strategies and memory-efficient mechanisms for incremental signature updates. Second, we plan to integrate ISC with other modern offline reinforcement learning objectives and frameworks, such as conservative or diffusion-based policies, and evaluate their performance on larger-scale, domain-specific benchmark datasets prior to real-world deployment. Finally, while ISCT currently employs a simple MLP-based decoding scheme for ISC tokens, exploring more expressive architectures for decoding incremental signature representations remains an important direction for further improving performance.

More broadly, this work highlights the importance of temporally structured representations in sequential decision learning. By bridging path signature theory with modern sequence models, ISC provides a principled and flexible framework for incorporating rich historical information into offline reinforcement learning, offering a promising direction for future research at the intersection of representation learning and control.

\bibliographystyle{IEEEtran}
\bibliography{refs}

@inproceedings{fu2020d4rl,
  title     = {D4RL: Datasets for Deep Data-Driven Reinforcement Learning},
  author    = {Justin Fu and Aviral Kumar and Ofir Nachum and George Tucker and Sergey Levine},
  year      = {2021},
  url       = {https://arxiv.org/abs/2004.07219},
  eprint    = {2004.07219},
  primaryClass = {cs.LG},
  archivePrefix = {arXiv}
}

@article{lyons1998roughsignals,
  title   = {Differential Equations Driven by Rough Signals},
  author  = {Lyons, Terry J.},
  journal = {Revista Matem{\'a}tica Iberoamericana},
  volume  = {14},
  number  = {2},
  pages   = {215--310},
  year    = {1998},
  url     = {https://eudml.org/doc/39555}
}

@book{lyonscaruana2007roughpaths,
  title     = {Differential Equations Driven by Rough Paths},
  author    = {Lyons, Terry and Caruana, Michael and L{\'e}vy, Thierry},
  series    = {Lecture Notes in Mathematics},
  volume    = {1908},
  publisher = {Springer Berlin, Heidelberg},
  year      = {2007},
  url       = {https://link.springer.com/book/10.1007/978-3-540-71285-5}
}

@article{levin2013learning,
  title   = {Learning from the past, predicting the statistics for the future, learning an evolving system},
  author  = {Daniel Levin and Terry Lyons and Hao Ni},
  year    = {2013},
  eprint  = {1309.0260},
  archivePrefix = {arXiv},
  primaryClass={q-fin.ST},
  url     = {https://arxiv.org/abs/1309.0260}
}

@article{graham2013sparse,
  title   = {Sparse arrays of signatures for online character recognition},
  author  = {Benjamin Graham},
  year    = {2013},
  eprint  = {1308.0371},
  archivePrefix = {arXiv},
  primaryClass={cs.CV},
  url     = {https://arxiv.org/abs/1308.0371}
}

@inproceedings{morrill2021neuralrde,
  title     = {Neural Rough Differential Equations for Long Time Series},
  author    = {James Morrill and Cristopher Salvi and Patrick Kidger and James Foster and Terry Lyons},
  year      = {2021},
  url       = {https://arxiv.org/abs/2009.08295},
  eprint    = {2009.08295},
  primaryClass={cs.LG},
  archivePrefix = {arXiv}
}

@inproceedings{kumar2020cql,
  title     = {Conservative Q-Learning for Offline Reinforcement Learning},
  author    = {Kumar, Aviral and Zhou, Aurick and Tucker, George and Levine, Sergey},
  year      = {2020},
  publisher = {Curran Associates Inc.},
  url       = {https://dl.acm.org/doi/abs/10.5555/3495724.3495824},
  booktitle = {Proceedings of the 34th International Conference on Neural Information Processing Systems}
}

@inproceedings{fujimoto2021td3bc,
  title     = {A Minimalist Approach to Offline Reinforcement Learning},
  author    = {Fujimoto, Scott and Gu, Shixiang Shane},
  publisher = {Curran Associates Inc.},
  year      = {2021},
  url       = {https://dl.acm.org/doi/10.5555/3540261.3541801},
  booktitle = {Proceedings of the 35th International Conference on Neural Information Processing Systems},
  articleno = {1540},
  series = {NIPS '21}
}

@inproceedings{
  kostrikov2022iql,
  title={Offline Reinforcement Learning with Implicit Q-Learning},
  author={Ilya Kostrikov and Ashvin Nair and Sergey Levine},
  booktitle={International Conference on Learning Representations},
  year={2022},
  url={https://openreview.net/forum?id=68n2s9ZJWF8}
}

@inproceedings{chen2021dt,
  author = {Chen, Lili and Lu, Kevin and Rajeswaran, Aravind and Lee, Kimin and Grover, Aditya and Laskin, Michael and Abbeel, Pieter and Srinivas, Aravind and Mordatch, Igor},
  title = {Decision transformer: reinforcement learning via sequence modeling},
  year = {2021},
  isbn = {9781713845393},
  publisher = {Curran Associates Inc.},
  booktitle = {Proceedings of the 35th International Conference on Neural Information Processing Systems},
  articleno = {1156},
  numpages = {14},
  series = {NIPS '21},
  url= {https://dl.acm.org/doi/abs/10.5555/3540261.3541417}
}

@inproceedings{janner2021trajectorytransformer,
  author = {Janner, Michael and Li, Qiyang and Levine, Sergey},
  title = {Offline reinforcement learning as one big sequence modeling problem},
  year = {2021},
  isbn = {9781713845393},
  publisher = {Curran Associates Inc.},
  address = {Red Hook, NY, USA},
  booktitle = {Proceedings of the 35th International Conference on Neural Information Processing Systems},
  articleno = {98},
  numpages = {14},
  series = {NIPS '21},
  url = {https://dl.acm.org/doi/10.5555/3540261.3540359}
}

@inproceedings{
  wang2022diffusionql,
  title={Diffusion Policies as an Expressive Policy Class for Offline Reinforcement Learning},
  author={Zhendong Wang and Jonathan J Hunt and Mingyuan Zhou},
  booktitle={The Eleventh International Conference on Learning Representations },
  year={2023},
  url={https://openreview.net/forum?id=AHvFDPi-FA}
}

@article{chi2023diffusionpolicy,
  title   = {Diffusion Policy: Visuomotor Policy Learning via Action Diffusion},
  author  = {Cheng Chi and Zhenjia Xu and Siyuan Feng and Eric Cousineau and Yilun Du and Benjamin Burchfiel and Russ Tedrake and Shuran Song},
  year    = {2024},
  eprint  = {2303.04137},
  archivePrefix = {arXiv},
  primaryClass={cs.RO},
  url     = {https://arxiv.org/abs/2303.04137}
}

@inproceedings{ni22a,
  title = 	 {Recurrent Model-Free {RL} Can Be a Strong Baseline for Many {POMDP}s},
  author =       {Ni, Tianwei and Eysenbach, Benjamin and Salakhutdinov, Ruslan},
  booktitle = 	 {Proceedings of the 39th International Conference on Machine Learning},
  pages = 	 {16691--16723},
  year = 	 {2022},
  volume = 	 {162},
  series = 	 {Proceedings of Machine Learning Research},
  month = 	 {17--23 Jul},
  publisher =    {PMLR},
  url = 	 {https://proceedings.mlr.press/v162/ni22a.html},
}

@article{kidger2020ncde,
  author = {Kidger, Patrick and Morrill, James and Foster, James and Lyons, Terry},
  title = {Neural controlled differential equations for irregular time series},
  year = {2020},
  isbn = {9781713829546},
  publisher = {Curran Associates Inc.},
  address = {Red Hook, NY, USA},
  booktitle = {Proceedings of the 34th International Conference on Neural Information Processing Systems},
  articleno = {562},
  numpages = {12},
  location = {Vancouver, BC, Canada},
  series = {NIPS '20},
  url = {https://dl.acm.org/doi/10.5555/3495724.3496286}
}

@inproceedings{jhin2023learnablepath,
  title        = {Learnable Path in Neural Controlled Differential Equations},
  author       = {Jhin, Sheo Yon and Jo, Minju and Kook, Seungji and Park, Noseong},
  year         = {2023},
  booktitle = {Proceedings of the Thirty-Seventh AAAI Conference on Artificial Intelligence and Thirty-Fifth Conference on Innovative Applications of Artificial Intelligence and Thirteenth Symposium on Educational Advances in Artificial Intelligence},
  articleno = {900},
  doi = {10.1609/aaai.v37i7.25969},
  publisher = {AAAI Press},
  url          = {https://doi.org/10.1609/aaai.v37i7.25969}
}

@inproceedings{walker2024logncde,
  title        = {Log neural controlled differential equations: the lie brackets make a difference},
  author       = {Walker, Benjamin and McLeod, Andrew D. and Qin, Tiexin and Cheng, Yichuan and Li, Haoliang and Lyons, Terry},
  year         = {2024},
  publisher = {JMLR.org},
  booktitle = {Proceedings of the 41st International Conference on Machine Learning},
  articleno = {2037},
  url          = {https://dl.acm.org/doi/10.5555/3692070.3694107}
}

@inproceedings{tong2023sigformer,
  title     = {SigFormer: Signature Transformers for Deep Hedging},
  author    = {Tong, Anh and Nguyen-Tang, Thanh and Lee, Dongeun and Tran, Toan M and Choi, Jaesik},
  booktitle = {4th ACM International Conference on AI in Finance},
  publisher={ACM},
  year      = {2023},
  month=nov, pages={124–132}, 
  archivePrefix = {arXiv},
  DOI={10.1145/3604237.3626841},
  url       = {http://dx.doi.org/10.1145/3604237.3626841}
}

@inproceedings{
  moreno2024roughtransformer,
  title={Rough Transformers: Lightweight Continuous-Time Sequence Modelling with Path Signatures},
  author={Fernando Moreno-Pino and Alvaro Arroyo and Harrison Waldon and Xiaowen Dong and Alvaro Cartea},
  booktitle={The Thirty-eighth Annual Conference on Neural Information Processing Systems},
  year={2024},
  url={https://openreview.net/forum?id=gXWmhzeVmh}
}

@inproceedings{
  barancikova2024sigdiffusions,
  title={SigDiffusions: Score-Based Diffusion Models for Time Series via Log-Signature Embeddings},
  author={Barbora Barancikova and Zhuoyue Huang and Cristopher Salvi},
  booktitle={The Thirteenth International Conference on Learning Representations},
  year={2025},
  url={https://openreview.net/forum?id=Y8KK9kjgIK}
}

@inproceedings{zhao2024cmfil,
  title        = {Mean Field Correlated Imitation Learning},
  author       = {Zhao, Zhiyu and Ma, Chengdong and Mi, Qirui and Yang, Ning and Yan, Xue and Yang, Mengyue and Zhang, Haifeng and Wang, Jun and Yang, Yaodong},
  booktitle = {Proceedings of the 24th International Conference on Autonomous Agents and Multiagent Systems},
  year         = {2025},
  pages = {2364–2372},
  eprint={2404.09324},
  url          = {https://dl.acm.org/doi/10.5555/3709347.3743877}
}

@misc{yamagata2023qlearningdecisiontransformerleveraging,
      title={Q-learning Decision Transformer: Leveraging Dynamic Programming for Conditional Sequence Modelling in Offline RL}, 
      author={Taku Yamagata and Ahmed Khalil and Raul Santos-Rodriguez},
      year={2023},
      eprint={2209.03993},
      archivePrefix={arXiv},
      primaryClass={cs.LG},
      url={https://arxiv.org/abs/2209.03993}, 
}

@misc{lyons2025signaturemethodsmachinelearning,
      title={Signature Methods in Machine Learning}, 
      author={Terry Lyons and Andrew D. McLeod},
      year={2025},
      eprint={2206.14674},
      archivePrefix={arXiv},
      primaryClass={stat.ML},
      url={https://arxiv.org/abs/2206.14674}, 
}

@misc{furuta2022generalizeddecisiontransformeroffline,
      title={Generalized Decision Transformer for Offline Hindsight Information Matching}, 
      author={Hiroki Furuta and Yutaka Matsuo and Shixiang Shane Gu},
      year={2022},
      eprint={2111.10364},
      archivePrefix={arXiv},
      primaryClass={cs.LG},
      url={https://arxiv.org/abs/2111.10364}, 
}

@misc{zheng2022onlinedecisiontransformer,
      title={Online Decision Transformer}, 
      author={Qinqing Zheng and Amy Zhang and Aditya Grover},
      year={2022},
      eprint={2202.05607},
      archivePrefix={arXiv},
      primaryClass={cs.LG},
      url={https://arxiv.org/abs/2202.05607}, 
}

@misc{correia2022hierarchicaldecisiontransformer,
      title={Hierarchical Decision Transformer}, 
      author={André Correia and Luís A. Alexandre},
      year={2022},
      eprint={2209.10447},
      archivePrefix={arXiv},
      primaryClass={cs.LG},
      url={https://arxiv.org/abs/2209.10447}, 
}
\vspace{12pt}

\appendix

For the group division, we use the following 3 groups: $\{2,3,4,5,6,7\},\,\{11,12,13,14,15,16\},\, \{0,1,8,9,10\}$ for halfcheetah and walker2d; we use the following 2 groups: $\{0,1,2,3,4\},\,\{5,6,7,8,9,10\}$ for hopper. we use the following 2 groups: $\{0,1\},\,\{2,3\}$ for maze2d. This channel separation is based on the real physical meaning of the state dimensions. Other hyperparameters we used are displayed in the table below.

The transformer architecture used in this paper is minGPT implemented by Andrej Karpathy.

\begin{table}[h]
\centering
\begin{tabular}{@{}l l@{}}
\toprule
\textbf{Hyperparameter} & \textbf{Value} \\
\midrule
Number of layers & 4 \\
Number of attention heads & 4 \\
Embedding dimension & 128 \\
Nonlinearity function & GeLU \\
Batch size & 256 \\
Truncated level & 2\\
Context length $K$ &
\makecell[l]{50 HalfCheetah, Hopper, Walker2d\\
20 Maze2d} \\
Goal conditioning &
\makecell[l]{300 HalfCheetah\\
150 Hopper\\
250 Walker2d\\
30 Maze2d\\} \\

Dropout & 0.1 \\
Learning rate & $10^{-3}$ \\
Grad norm clip & $1$ \\
Weight decay & $0.01$ \\
Learning linear warmup & first $10$ Epochs  \\
\bottomrule
\end{tabular}
\end{table}

\end{document}